\documentclass{article}

\usepackage{microtype}
\usepackage{graphicx}
\usepackage{subfigure}
\usepackage{booktabs} % for professional tables
\usepackage{amsmath}
\usepackage{caption}
\usepackage{comment}
\usepackage{quoting}
\usepackage{tikz}
\usetikzlibrary{calc}
\usepackage{mwe}
\usetikzlibrary{positioning}
\usepackage{color, xcolor}

\usepackage{makecell}
\usepackage{siunitx}
\DeclareMathOperator{\arctanh}{arctanh}\usepackage{amsthm,amsfonts}
\usepackage[pagestyles]{titlesec}
\usepackage{fancyhdr}

\newcommand{\alert}[1]{\textbf{}}
\newcommand{\red}[1]{\textbf{}}

\newcommand{\BEAS}{\begin{eqnarray*}}
\newcommand{\EEAS}{\end{eqnarray*}}
\newcommand{\BEA}{\begin{eqnarray}}
\newcommand{\EEA}{\end{eqnarray}}
\newcommand{\BEQ}{\begin{equation}}
\newcommand{\EEQ}{\end{equation}}
\newcommand{\BIT}{\begin{itemize}}
\newcommand{\EIT}{\end{itemize}}
\newcommand{\BNUM}{\begin{enumerate}}
\newcommand{\ENUM}{\end{enumerate}}
\newcommand{\BEL}[1]{\begin{equation}\label{#1}}
\newcommand{\EEL}{\end{equation}}

\newcommand{\BA}{\begin{array}}
\newcommand{\EA}{\end{array}}

\newcommand{\eg}{{\it e.g.}}
\newcommand{\ie}{{\it i.e.}}

\newcommand{\ones}{\mathbf 1}

\newcommand{\reals}{\mathbb{R}}

\newcommand{\Tr}{\mathop{\bf Tr}}
\newcommand{\diag}{\mathop{\bf diag}}

\newcommand{\dsp}{\displaystyle}

\newtheorem{theorem}{Theorem}

\newtheorem{remark}[theorem]{Remark}

\newcounter{exno}

\makeatletter
\newcounter{tablecounter}

\makeatother

\usepackage{hyperref}

\usepackage[accepted]{icml2018}

\icmltitlerunning{Lifted Neural Networks}

\begin{document}

\twocolumn[
\icmltitle{Lifted Neural Networks}



\icmlsetsymbol{equal}{*}

\begin{icmlauthorlist}
\icmlauthor{Armin Askari}{eecs}
\icmlauthor{Geoffrey Negiar}{eecs}
\icmlauthor{Rajiv Sambharya}{eecs}
\icmlauthor{Laurent El Ghaoui}{eecs,ieor}
\end{icmlauthorlist}

\icmlaffiliation{eecs}{Department of Electrical Engineering and Computer Science, University of California --- Berkeley, United States}
\icmlaffiliation{ieor}{Department of Industrial Engineering and Operations Research, University of California --- Berkeley, United States}

%\icmlcorrespondingauthor{Armin Askari}{aaskari@berkeley.edu}
\icmlcorrespondingauthor{Armin Askari}{aaskari@berkeley.edu}
\icmlcorrespondingauthor{Geoffrey Negiar}{geoffrey\_negiar@berkeley.edu}

\icmlkeywords{Machine Learning, ICML}

\vskip 0.3in
]



\printAffiliationsAndNotice{\icmlEqualContribution} % otherwise use the standard text.

\begin{abstract}
 We describe a novel family of models of multi-layer feedforward neural networks in which the activation functions are encoded via penalties in the training problem. Our approach is based on representing  a non-decreasing activation function as the argmin of an appropriate convex optimization problem. The new framework allows for algorithms such as block-coordinate descent methods to be applied, in which each step is composed of a simple (no hidden layer) supervised learning problem that is parallelizable across data points and/or layers. Experiments indicate that the proposed models provide excellent initial guesses for weights for standard neural networks. In addition, the model provides avenues for interesting extensions, such as robustness against noisy inputs and optimizing over parameters in activation functions.
\end{abstract}

\section{Introduction} % (fold)
\label{sec:introduction}

Given current advances in computing power, dataset sizes and the availability of specialized hardware/ software packages, the popularity of neural networks continue to grow. The model has become standard in a large number of tasks, such as image recognition, image captioning and machine translation. Current state of the art is to train this model by variations of stochastic gradient descent (SGD), although these methods have several caveats. Most problems with SGD are discussed in \cite{Taylor:2016:TNN:3045390.3045677ADMM}.

Optimization methods for neural networks has been an active research topic in the last decade. Specialized gradient-based algorithms such as Adam and \textsc{Adagrad} \citep{KingmaB14Adam, DBLP:journals/jmlr/DuchiHS11} are often used but were shown to generalize less than their non adaptive counterparts by \cite{DBLP:conf/nips/WilsonRSSR17Marginal}. Our work is related to two main currents of research aimed at improving neural network optimization: using non gradient-based approaches and initializing weights to accelerate convergence of gradient-based algorithms. To our knowledge this paper is the first to combine the two. In addition our novel formalism allow for interesting extensions towards handling constraints, robustness, optimizing network topology, etc.

\citep{Taylor:2016:TNN:3045390.3045677ADMM} and \cite{pmlr-v33-carreira-perpinan14NestedSystems} propose an approach similar to ours, adding variables in the training problem and using an $l^2$-norm penalization of equality constraints. They both break down the network training problem into easier sub-problems and use alternate minimization; however they do not exploit structure in the activation functions. For Convolutional Neural Networks (CNN), \citep{Berrada2016TrustingSF} model the network training problem as a difference of convex functions optimization, where each subproblem is a Support Vector Machine (SVM). 

On the initialization side, \cite{LeCun:1998:EB:645754.668382, Glorot10understandingthe} recommend sampling from a well-chosen uniform distribution to initialize weights and biases while others either use random initialization or weights learned in other networks (transfer learning) on different tasks. \cite{Sutskever:2013:IIM:3042817.3043064} indicate that initialization is crucial during training and that poorly initialized networks cannot be trained with momentum. Other methods to initialize neural networks have been proposed, such as using competitive learning \cite{Maclin:1995:CPM:1625855.1625924} and principal component analysis (PCA) \cite{DBLP:journals/corr/SeuretAIL17}. Although PCA produces state of the art results, it is limited to auto-encoders while our framework allows for more general learning problems. Similarly, the competitive learning approach is limited to the classification problem and works only for one layer networks while our model can easily be adapted to a broader range of network architectures. Our approach focuses on transforming the non-smooth optimization problem encountered when fitting neural network models into a smooth problem in an enlarged space; this ties to a well developed branch of optimization literature (see \eg \ section $5.2$ of \citep{Bubeck:2015:COA:2858997.2858998} and references therein). Our approach can also be seen as a generalization of the parameterized rectified linear unit (PReLU) proposed by \citep{DBLP:journals/corr/HeZR015prelu}. Our work can be compared to the standard practice of initializing Gaussian Mixture Models using $K$-Means clustering; our model uses a simpler but similar algorithm for initialization.

\textbf{Paper outline.} In Section~\ref{sec:background}, we begin by describing the mathematical setting of neural networks and our proposed optimization problem to train the model. Section~\ref{sec:basic_idea} provides an example illustrating the basic idea. Section~\ref{sec:activation_as_opt} outlines how to encode activation functions as argmins of convex or bi-convex optimization problems. Section~\ref{sec:lifted_framework} then expands the approach of Section~\ref{sec:basic_idea} to cover a number of useful activation functions, as well as classification tasks. Section~\ref{sec:block_coordinate_descent_approach} describes a block-coordinate descent method to solve the training problem. Section~\ref{sec:numerical_experiments} describes numerical experiments that support a finding that the models can be used as a fast weight initialization scheme.
% section introduction (end)

\section{Background and Notation}
\label{sec:background}
\paragraph{Feedforward neural networks.} % (fold)
\label{par:feedforward_neural_networks}
We begin by establishing notation. We are given an input data matrix $X = [x_1,\ldots,x_m] \in \reals^{n \times m}$ and response matrix $Y \in \reals^{p\times m}$ and consider a supervised problem involving a neural network having $L \ge 1$ hidden layers. At test time, the network processes an input vector $x \in \reals^n$ to produce a predicted value $\hat{y}(x) \in \reals^p$ according to the prediction rule $\hat{y}(x) = x_{L+1}$
% \begin{equation}\label{eq:std-pred-rule}
% \hat{y}(x) = \phi_L(W_Lx_L + b_L) %\arg \min_Y \: {\cal L}(Y,W_Lx_L + b_L),
% \end{equation}
where $x_{L+1}$ is defined via the recursion
\begin{equation}
\label{eq:recursion-x}
x_{l+1} = \phi_l(W_lx_l + b_l), \;\; l=0,\ldots,L,
\end{equation}
with initial value $x_0 = x \in \reals^n$ and $x_l \in \reals^{p_l}$, $l=0,\ldots,L$.  Here, $\phi_l$, $l=1,\ldots,L$ are given activation functions, acting on a vector; the matrices $W_l \in \reals^{p_{l+1}\times p_l}$ and vectors $b_l \in \reals^{p_{l+1}}$, $l=0,\ldots,L$ are parameters of the network. In our setup, the sizes $(p_l)_{l=0}^{L+1}$ are given with $p_0 = n$ (the dimension of the input) and $p_{L+1} = p$ (the dimension of the output). 

We can express the predicted outputs for a given set of $m$ data points contained in the $n \times m$ matrix $X$ as the $p \times m$ matrix $\hat{Y}(X) = X_{L+1}$, as defined by the matrix recursion
\begin{equation}\label{eq:recursion-nn}
X_{l+1} = \phi_l(W_lX_l +b_l\ones^T), \;\; l=0,\ldots,L,
\end{equation}
with initial value $X_0 = X$ and $X_l \in \reals^{p_l \times m}$, $l=0,\ldots,L$. Here, $\ones$ stands for the vector of ones in $\reals^m$, and we use the convention that the activation functions act column-wise on a matrix input.

In a standard neural network, the matrix parameters of the network are fitted via an optimization problem, typically of the form
\begin{equation}\label{eq:penalized-nn-standard}
\begin{array}{rl}
& \dsp\min_{(W_l,b_l)_{0}^{L},(X_l)_{1}^{L}} 
 {\cal L}(Y,X_{L+1})
+ \dsp\sum_{l=0}^{L} \rho_l \pi_l(W_l)  \\
\mbox{s.t.} & X_{l+1} = \phi_l(W_lX_l +b_l\ones_m^T), \; \; l=0,\ldots,L \\
& X_0 = X
\end{array}
\end{equation}
where ${\cal L}$ is a loss function, $\rho \in \reals_{+}^{L+1}$ is a hyper-parameter vector, and $\pi_l$'s are penalty functions which can be used to encode convex constraints, network structure, etc. We refer to the collections $(W_l,b_l)_{l=0}^{L}$ and $(X_l)_{l=1}^{L}$ as the $(W,b)$- and $X$-variables, respectively.  

To solve the training problem~(\ref{eq:penalized-nn-standard}), the $X$-variables are usually eliminated via the recursion~(\ref{eq:recursion-nn}), and the resulting objective function of the $(W,b)$-variables is minimized without constraints, via stochastic gradients. While this appears to be a natural approach, it does make the objective function of the problem very complicated and difficult to minimize.
% paragraph feedforward_neural_networks (end)

\textbf{Lifted models.} % (fold)
\label{par:lifted_models}
In this paper, we develop a family of models where the $X$-variables are kept, and the recursion constraints~(\ref{eq:recursion-x}) are approximated instead, via penalties. We refer to these models as ``lifted'' because we lift the search space of $(W,b)$-variables to a higher-dimensional space of $(W,b,X)$-variables.  The training problem is cast in the form of a matrix factorization problem with constraints on the variables encoding network structure and activation functions.

Lifted models have many more variables but a much more explicit structure than the original, allowing for training algorithms that can use efficient standard machine learning libraries in key steps. The block-coordinate descent algorithm described here involves steps that are parallelizable across either data points and/or layers; each step is a simple structured convex problem.

The family of alternate models proposed here have the potential to become competitive in their own right in learning tasks, both in terms of speed and performance. In addition, such models are versatile enough to tackle problems deemed difficult in a standard setting, including robustness to noisy inputs, adaptation of activation functions to data, or including constraints on the weight matrices. Our preliminary experiments are limited to the case where the lifted model's variables are used as initialization of traditional feedforward network. However, we discuss and layout the framework for how these models can be used to tackle other issues concerning traditional networks such as robustness and optimizing how to choose activation functions at each layer. %These experiments show that such an initialization scheme results in improved performance, and improved convergence speed.
% paragraph lifted_models (end)

\section{Basic Idea} % (fold)
\label{sec:basic_idea}
To describe the basic idea, we consider a specific example, in which all the activation functions are the ReLUs, except for the last layer. There $\phi_L$ is the identity for regression tasks or a softmax for classification tasks. In addition, we assume in this section that the penalty functions are of the form $\pi_l(W) = \|W\|_F^2$, $l=0,\ldots,L$.

We observe that the ReLU map, acting componentwise on a vector input $u$, can be represented as the ``argmin'' of an optimization problem involving a jointly convex function:
\begin{equation}\label{eq:relu-rep}
\phi(u) = \max(0,u) = \arg\min_{v \ge 0} \: \|v-u\|_2.
\end{equation}
As seen later, many activation functions can be represented as the ``$\arg\min$'' of an optimization problem, involving a jointly convex or bi-convex function.  

Extending the above to a matrix case yields that the condition $X_{l+1} = \phi(W_lX_l+b_l\ones^T)$ for given $l$ can be expressed via an ``$\arg\min$'':
\[
X_{l+1} \in \arg\min_{Z\geq 0} \: \|Z-W_lX_l-b_l\ones^T\|_F^2.
\]
This representation suggests a heuristic to solve~(\ref{eq:penalized-nn-standard}), replacing the training problem by
\begin{equation}\label{eq:lifted-ridge-lambdas}
\begin{array}{rl}
\dsp\min_{(W_l,b_l),(X_l)} & 
\mathcal{L}(Y,W_LX_L + b_L\textbf{1}^T) + \dsp\sum_{l=0}^L \rho_l \|W_l\|_F^2 \\
 & + \dsp\sum_{l=0}^{L-1}  \left( \lambda_{l+1} \|X_{l+1}-W_lX_l-b_l\ones^T\|_F^2 \right) \\
\mbox{s.t.} & X_l \ge 0, \;\; l=1,\ldots,L-1, \;\;
X_0 = X.
\end{array}
\end{equation}
where $\lambda_{l+1}>0$ are hyperparameters, $\rho_l$ are regularization parameters as in (\ref{eq:penalized-nn-standard}) and $\mathcal{L}$ is a loss describing the learning task. In the above model, the activation function is not used in a pre-defined manner; rather, it is \emph{adapted} to data, via the non-negativity constraints on the ``state'' matrices $(X_l)_{l=1}^{L+1}$.  We refer to the above as a ``lifted neural network'' problem. 

Thanks to re-scaling the variables with $X_l \rightarrow \sqrt{\lambda_l} X_l$, $W_l \rightarrow \sqrt{\lambda_{l+1}/\lambda_l} W_l$, and modifying $\rho_l$'s accordingly, we can always assume that all entries in $\lambda$ are equal, which means that our model introduces just one extra scalar hyper-parameter over the standard network~(\ref{eq:penalized-nn-standard}).

The above optimization problem is, of course, challenging, mainly due to the number of variables. However, for that price we gain a lot of insight on the training problem. In particular, the new model has the following useful characteristics:
\begin{itemize}
\item For fixed $(W,b)$-variables, the problem is convex in the $X$-variables $X_l$, $l=1,\ldots,L$; more precisely it is a (matrix) non-negative least-squares problem. The problem is fully \emph{parallelizable across the data points}.
\item Likewise, for fixed $X$-variables, the problem is convex in the $(W,b)$-variables and \emph{parallelizable across layers and data points}. In fact, the $(W,b)$-step is a set of parallel (matrix) ridge regression problems.
\end{itemize}
These characteristics allow for efficient block-coordinate descent methods to be applied to our learning problem. Each step reduces to a basic supervised learning problem, such as ridge regression or non-negative least-squares. We describe one algorithm in more detail in section~\ref{sec:block_coordinate_descent_approach}. 

The reader may wonder at this point what is the prediction rule associated with our model. For now, we focus on extending the approach to broader classes of activations and loss functions used in the last layer; we return to the prediction rule issue in our more general setting in section~\ref{sub:prediction_rule}.
% subsection basic_idea (end)

\section{Activations as $\arg\min$ Maps} % (fold)
\label{sec:activation_as_opt}

In this section, we outline theory on how to convert a class of functions as the ``$\arg\min$'' of a certain optimization problem, which we then encode as a penalty in the training problem. We make the following assumption on a generic activation function $\phi$. 
\begin{quoting}[vskip=1pt]
\begin{quote} 
\textbf{BCR Condition.} The activation function $\phi : \reals^k \rightarrow \reals^j$ satisfies the bi-convex representation (BCR) condition if it can be represented as follows:
\[
\forall x\in \reals^k,\; \phi(x) = \arg\min_{z \in \reals^j} \: {\cal D}_\phi(x,z)  ,
\]
where ${\cal D}_\phi : \reals^k \times \reals^j \rightarrow \reals$ is a bi-convex function (convex in $x$ for fixed $z$ and vice-versa), which is referred to as a \emph{BC-divergence} associated with the activation function.
\end{quote}
\end{quoting}
We next examine a few examples, all based on divergences of the form
\[
D_\phi(x,z) = \Phi (z) - x^T z ,
\]
where $\Phi$ is a convex function. This form implies that, when $\Phi$ is differentiable, $\phi$ is the gradient map of a convex function; thus, it is monotone.

\paragraph{Strictly monotone activation functions.} % (fold)
\label{par:monotone_activation_functions}
We assume that $\phi$ is strictly monotone, say without loss of generality, strictly increasing. Then, it is invertible, and there exists a function, denoted $\phi^{-1}$, such that the condition $x = \phi^{-1}(z)$ for $z \in \mbox{\bf range}(\phi)$ implies $z = \phi(x)$.  Note that $\phi^{-1}$ is strictly increasing on its domain, which is $\mbox{\bf range}(\phi)$.

Define the function $\Phi : \reals \rightarrow \reals$, with values
\begin{equation}\label{eq:Phi-def}
\Phi(z) = \int_0^z \phi^{-1}(u) \: du \mbox{ if } z \in \mbox{\bf range}(\phi),
\end{equation}
and $+\infty$ otherwise.

The function $\Phi$ is convex, since $\phi^{-1}$ is increasing. We then consider the problem
\begin{equation}\label{eq:basicpb}
\min\{ \: \Phi(z) - xz ~:~ z \in \mbox{\bf range}(\phi)\}.
\end{equation}
Note that the \emph{value} of the problem is nothing else than $\Phi^*(x)$, where $\Phi^*$ is the Fenchel conjugate of $\Phi$.

By construction, the problem~(\ref{eq:basicpb}) is convex. At optimum, we have $x = \phi^{-1}(z)$, hence $z = \phi(x)$. We have obtained
\[
\phi(x) = \arg\min_{z} \: \Phi(z) - xz ~:~ z \in \mbox{\bf range}(\phi).
\]
% paragraph monotone_activation_functions (end)

\paragraph{Examples.} % (fold)
\label{par:examples}
As an example, consider the sigmo\"id function:
\[
\phi(x) = \frac{1}{1+e^{-x}},
\]
with inverse 
\[
\phi^{-1}(z) = \log\frac{z}{1-z} , \;\; 0 < z < 1,
\]
and $+\infty$ otherwise.

Via the representation result~(\ref{eq:Phi-def}), we obtain
\[
\phi(x) = \arg\min_{0 \leq z \leq 1} \: z\log z + (1-z) \log (1-z) - xz
\]

Next consider the ``leaky ReLU'' function
\[
\phi(x) = \left\{ \begin{array}{ll} \alpha x & \mbox{if } x < 0, \\ x & \mbox{if } x \ge 0,
\end{array} \right.
\]
where $0 <\alpha <1$. We have
\[
\phi^{-1}(z) = \left\{ \begin{array}{ll} (1/\alpha) z & \mbox{if } z < 0, \\ z & \mbox{if } z \ge 0,
\end{array} \right.
\]
with domain the full real line; thus
\begin{align}\label{eq:def-Phi-leaky}
\Phi(z) &= \int_0^z \phi^{-1}(u) \: du \nonumber \\
&= \frac{1}{2}\max \left( 
 \frac{1}{\alpha}\max(0,-z)^2 , \max(0,z)^2 \right)
\end{align}
As another example, consider the case with $\phi(x) = \arctanh (x)$. The inverse function is 
\[
\phi^{-1}(z) = \frac{1}{2} \log \frac{1+z}{1-z} , \;\; |z| \le 1
\]
For any $z \in [-1,1]$, $\Phi(z)$ takes the form 
\begin{align*}
\Phi(z) &=  \frac{1}{2} \int_0^z \left( \log (1+u) - \log(1-u) \right) \: du \\
&= \frac{1}{2} \left( (1-z) \log (1-z) + (1+z) \log(1+z) \right) + \mbox{cst.}
\end{align*}

Sometimes there are no closed-form expressions. For example, for the so-called ``softplus'' function $\phi(x) = \log(1+e^x)$, the function $\Phi$ cannot be expressed in closed form:
\[
\Phi(z) = \int_0^z \log(e^u - 1) \: du, \;\; \mbox{\bf dom} \Phi = \reals_+.
\]
This lack of a closed-form expression does not preclude algorithms from work with these types of activation functions. The same is true of the sigmoid function.
% paragraph examples (end)

\paragraph{Non-strictly monotone examples: ReLU and piece-wise sigmoid.} % (fold)
\label{par:a_non_invertible_example}
The above expression~(\ref{eq:basicpb}) works in the ReLU case; we simply restrict the inverse function to the domain $\reals_+$; specifically, we define
\[
\phi^{-1} (z) = \left\{ \begin{array}{ll} +\infty & \mbox{if } z < 0, \\ z & \mbox{if } z \ge 0,
\end{array} \right.
\]
We then have $\mbox{\bf dom} \Phi = \reals_+$, and for $z \ge 0$:
\[
\Phi (z) = \int_0^z u \: du = \frac{1}{2} z^2.
\]
We have obtained
\[
\phi(x) = \arg\min_{z \ge 0} \: \Phi(z) - xz .
\]

The result is consistent with the ``leaky'' ReLU case in the limit when $\alpha \rightarrow 0$. Indeed, in that case with $\Phi$ given as in (\ref{eq:def-Phi-leaky}), we observe that when $\alpha \rightarrow 0$ the domain of $\Phi$ collapses from the whole real line to $\reals_+$, and the result follows.

In a similar vein, consider the ``piecewise'' sigmoid function, 
\[
\phi(x) = \min(1,\max(-1,x)),
\]
This function can be represented as 
\[
\phi(x) =\arg\min_{z} \: z^2 - 2xz ~:~ |z| \le 1.
\]
Finally the sign function is represented as
\[
\mbox{\bf sign}(x) = \arg\min \: -zx ~:~ |z| \le 1.
\]
% paragraph a_non_invertible_example (end)

\paragraph{Non-monotone examples.} % (fold)
\label{par:a_non_monotonic_example}
The approach can be sometimes extended to non-monotonic activation functions. As an example, the activation function $\phi(x) = \sin x$ has been proposed in the context of time-series. Here, we will work with
\[
\Phi(z) = \int_{0}^z \arcsin (u) \: du =  z \arcsin z + \sqrt{1-z^2} + \mbox{cst.},
\]
with domain $[-1,1]$. The function is convex, and we can check that
\[
\phi(x) = \arg\min_{z \::\: |z| \le 1} \: \Phi(x) - xz
\]

\paragraph{Jointly convex representations.} Some activation functions enjoy a stronger condition, which in turn leads to improved properties of the corresponding lifted model.
\begin{quote} 
\textbf{JCR Condition.} The activation function $\phi : \reals^k \rightarrow \reals^j$ satisfies the jointly convex representation (JCR) condition if it satisfies the CR condition with a jointly convex function ${\cal D}_\phi(x,z)$.
\end{quote}
Note that, for the JCR condition to hold, the activation function needs to be monotone. Because of non-uniqueness, we may add a term that is not a function of the variable being optimized (i.e. in the condition above, an arbitrary function of $u$) to the JC-divergence in order to improve the overall structure of the problem. This is highlighted below and discussed in Section \ref{sec:block_coordinate_descent_approach}.

The JCR condition applies to several important activation functions, beyond the ReLU, for which 
\[
\max(x,0) = \arg\min_z \: {\cal D}_\phi (x,z) = \left\{ \begin{array}{ll} \|x-z\|_2^2 & \mbox{if } z \ge 0, \\ +\infty & \mbox{otherwise.}
\end{array}
\right.
\]
Note that the JC-divergence for the ReLU is not unique; for example, we can replace the $l_2$-norm by the $l_1$-norm.

The ``leaky'' ReLU with parameter $\alpha \in (0,1)$, defined by $\phi(x) = \max(x/\alpha,x)$, can be written in a similar way:
\[
\max(x/\alpha,x) = \arg\min_z \: \|x-z\|_2^2 ~:~ z \ge (1/\alpha) x.
\]
The piece-wise sigmo\"id, as defined below, has a similar variational representation: with $\ones$ the vector of ones,
\[
\min(1,\max(0,x)) = \arg\min_z \: \|x-z\|_2^2 ~:~ 0 \le z \le \ones.
\]

In order to address multi-class classification problems, it is useful to consider a last layer with an activation function that produces a probability distribution output. To this end, we may consider an activation function which projects, with respect to some metric, a vector onto the probability simplex. The simplest example is the Euclidean projection of a real vector $u \in \reals^k$ onto the probability simplex in $\reals^k$:
\[
\phi(x) = \arg\min_z \: \|x-z\|_2^2 ~:~ z \ge 0, \;\; z^T\ones = 1.
\]

Max-pooling operators are often used in the context of image classification. A simple example of a max-pooling operator involves a $p$-vector input $x$ with two blocks, $x=(x^{(1)},x^{(2)})$, with $x{(i)}\in \reals^{p_i}$, $i=1,2$, with $p = p_1+p_2$. We define $\phi \::\: \reals^p \rightarrow \reals^2$ by
\begin{equation}\label{eq:max-pooling-2d}
\phi(x) = (\max_{1 \le i \le p_1} \: x^{(1)}_i,\max_{1 \le i \le p_2} \: x^{(2)}_i) \in \reals^2.
\end{equation}
Max-pooling operators can also be expressed in terms of a jointly convex divergence. 
In the above case, we have
\[
\phi(x) = \arg\min_z \: \ones^T z + \ones^T(x - Dz )_+,
\]
where $D$ is an appropriate block-diagonal matrix of size $p \times 2$ that encodes the specifics of the max-pooling, namely in our case $D = \diag(\ones_{p_1},\ones_{p_2})$.

\paragraph{Extension to matrix inputs.}
Equipped with a divergence function that works on vector inputs, we can readily extend it to matrix inputs with the convention that the divergence is summed across columns (data points). Specifically, if $X = [x_1,\ldots,x_m] \in \reals^{k \times m}$, we define  $\phi$ by $Z = \phi(X) = [z_1,\ldots,z_m] \in \reals^{h \times m}$ as acting column-wise. We have  
\[
\phi(X) := [\phi(x_1), \ldots, \phi(x_m)]= \arg\min_{Z} \:  {\cal D}_\phi(X,Z),
\]
where, with some minor abuse of notation, we define a matrix version of the divergence, as follows: 
\[
{\cal D}_\phi([x_1,\ldots,x_m],[z_1,\ldots,z_m]) = \sum_{i=1}^m {\cal D}_\phi(x_i,z_i).
\]

\section{Lifted Framework} % (fold)
\label{sec:lifted_framework}

\subsection{Lifted neural networks} % (fold)
\label{sub:lifted_neural_networks}
Assume that the BCR or JCR condition is satisfied for each layer of our network and use the short-hand notation $D_l = D_{\phi_l}$ for the corresponding divergences. Condition~(\ref{eq:recursion-nn}) is then written as
\[
X_{l+1} \in \arg\min_{X \in \mathcal{X}} \: D_l(X,W_lX_l+b_l\ones^T), \;\; l=0,\ldots,L.
\]
The lifted model consists in replacing the constraints~(\ref{eq:recursion-nn}) with penalties in the training problem. Specifically, the lifted network training problem takes the form
\begin{align}\label{eq:lifted-training}
\min_{(W_l,b_l),(X_l)} \: &{\cal L}(Y,W_LX_{L}+b_L\textbf{1}^T) + \sum_{l=0}^L \pi_l(W_l) \\ \nonumber
& + \sum_{l=0}^{L-1} \lambda_{l+1} D_l(W_lX_l+b_l\ones^T,X_{l+1})
\\ 
\nonumber
&\text{s.t. } X_0 = X, \; \; X_l \geq 0, \; l = 1,\hdots,L-1
\end{align}
with $\lambda_1,\ldots,\lambda_{L+1}$ given positive hyper-parameters.  As with the model introduced in section~\ref{sec:basic_idea}, the lifted model enjoys the same \textit{parallel and convex} structure outlined earlier. In particular, it is convex in $X$-variables for fixed $W$-variables. If we use a weaker bi-convex representation (using a bi-convex divergence instead of a jointly convex one), then convexity with respect to $X$-variables is lost. However, the model is still convex in $X_l$ for a given $l$ when all the other variables are fixed; this still allows for block-coordinate descent algorithms to be used.

As a specific example, consider a multi-class classification problem where all the layers involve ReLUs except for the last. The last layer aims at producing a probability distribution to be compared against training labels via a cross entropy loss function. The training problem writes
\begin{align}\label{eq:example-lifted}
\dsp\min_{(W_l,b_l),(X_l)} & -\Tr Y^T \log s(W_LX_L + b_L\textbf{1}^T) + \dsp\sum_{l=0}^{L} \rho_l \|W_l\|_F^2  \nonumber \\ 
&+ \dsp\sum_{l=0}^{L-1}   \lambda_{l+1} \|X_{l+1}-W_lX_l-b_l\ones^T\|_F^2 \nonumber \\
 &\mbox{s.t.} \;\;  X_0 = X, \; \;  X_l \ge 0, \;\; l=1,\hdots,L-1
\end{align}
where the equality constraint on $X_{L+1}$ enforces that its columns are probability distributions. Here,  $s(\cdot): \mathbb{R}^n \mapsto \mathbb{R}^n$ is the softmax function. We can always rescale the variables so that in fact the number of additional hyper-parameters $\lambda_l$, $l=1,\ldots,L-1$, is reduced to just one.
% subsection lifted_neural_networks (end)

\subsection{Lifted prediction rule} % (fold)
\label{sub:prediction_rule}
In our model, the prediction rule will be different from that of a standard neural network, but it is based on the same principle. In a standard network, the prediction rule can be obtained by solving the problem
\[
\hat{y}(x) = \min_y \: {\cal L}(y,x_{L+1}) 
~:~ (\ref{eq:recursion-nn}), \;\; x_0 = x,
\]
where the weights are now \emph{fixed}, and $y \in \reals^p$ is a \emph{variable}. Of course, provided the loss is zero whenever its two arguments coincide, the above trivially reduces to the standard prediction rule: $\hat{y}(x) = x_{L+1}$, where $x_{L+1}$ is obtained via the recursion~(\ref{eq:recursion-nn}). 

In a lifted framework, we use the same principle: solve the training problem (in our case, (\ref{eq:lifted-training})), using the test point as input, fixing the weights, and letting the predicted output values be variables. In other words, the prediction rule for a given test point $x$ in lifted networks is based on solving the problem
\begin{align}\label{eq:lifted-pre-rule}
\hat{y} = \arg&\min_{y,(x_l)} \: {\cal L}(y,W_Lx_{L} + b_L) \nonumber \\
&+ \sum_{l=0}^{L-1} \lambda_{l+1} D_l(W_lx_l+b_l,x_{l+1}) 
\nonumber
\\
&\text{s.t. }x_0 = x.
\end{align}
The above prediction rule is a simple convex problem in the variables $y$ and $x_l$, $l=1,\ldots,L$. In our experiments, we have found that applying the standard feedforward rule of traditional networks is often enough.
% subsection prediction_rule (end)

\section{Block-Coordinate Descent Algorithm} % (fold)
\label{sec:block_coordinate_descent_approach}
In this section, we outline a block-coordinate descent approach to solve the training problem~(\ref{eq:lifted-training}). 

\subsection{Updating $(W,b)$-variables} % (fold)
\label{sub:updating_w_variables}
For fixed $X$-variables, the problem of updating the $W$-variables, \ie\ the weighting matrices $(W_l,b_l)_{l=0}^L$, is parallelizable across both data points and layers. The sub-problem involving updating the weights at a given layer $l=0,\ldots,L$ takes the form
\[
(W_l^+,b_l^+) = \arg\min_{W,b} \: \lambda_{l+1} D_l(WX_l + b\ones^T,X_{l+1}) + \pi_l(W).
\]
The above is a convex problem, which can be solved via standard machine learning libraries. Since the divergences are sums across columns (data points), the above problem is indeed parallelizable across data points.

For example, when the activation function at layer $l$ is a ReLU, and the penalty $\pi_l$ is a squared Frobenius norm, the above problem reads
\[
(W_l^+,b_l^+) = \arg\min_{W,b} \: \lambda_{l+1} \|WX_l + b\ones^T - X_{l+1}\|_F^2 + \rho_l \|W\|_F^2
\]
which is a standard (matrix) ridge regression problem. Modern sketching techniques for high-dimensional least-squares can be employed, see for example~\cite{woodruff2014sketching,pilanci2016iterative}.

\subsection{Updating $X$-variables} % (fold)
\label{sub:updating_x_variables}
In this step we minimize over the matrices $(X_l)_{l=1}^{L+1}$. The sub-problem reads exactly as~(\ref{eq:lifted-training}), with now the $(W,b)$-variables fixed. By construction of divergences, the problem is decomposable across data points. When JCR conditions hold, the joint convexity of each JC-divergence function allows us update all the $X$-variables at once, by solving a convex problem. Otherwise, the update must be done cyclically over each layer, in a block-coordinate fashion. 

For $l = 1,\ldots,L$, the sub-problem involving $X_l$, with all the other $X$-variables $X_j$, $j \ne l$ fixed, takes the form
\begin{align}\label{eq:Xl-update-gen}
X_l^+ = \arg\min_{Z} \: &\lambda_{l+1} D_l(W_lZ+b_l\ones^T,X_{l+1}) \nonumber\\
+ & \lambda_{l} D_{l-1}(Z, X_{l-1}^0)
\end{align}
where $X_{l-1}^0 := W_{l-1}X_{l-1}+b_{l-1}\ones^T$. By construction, the above is a convex problem, and is again parallelizable across data points.

Let us detail this approach in the case when the layers $l,l+1$ are both activated by ReLUs. The sub-problem above becomes
\begin{align*}
X_l^+ = \arg\min_{Z \ge 0} \:& \lambda_{l+1} \|X_{l+1}- W_lZ - b_{l}\ones^T\|_F^2 + \\
& \lambda_{l}\|Z - W_{l-1}X_{l-1}-b_{l-1}\ones^T\|_F^2 
\end{align*}
The above is a (matrix) non-negative least-squares, for which many modern methods are available, see \cite{kim2007fast,kim2014algorithms} and references therein. As before, the problem above is fully parallelizable across data points (columns), where each data point gives rise to a standard (vector) non-linear least-squares. Note that the cost of updating all columns can be reduced by taking into account that all column's updates share the same coefficient matrix $W_l$.

The case of updating the last matrix $X_{L+1}$ is different, as it involves the output and the loss function ${\cal L}$. The update rule for $X_{L+1}$ is indeed
\begin{equation}\label{eq:Xlast-update-gen}
X_{L+1}^+ = \arg\min_{Z} \: {\cal L}(Y,Z) + \lambda_{L+1} D_L(X_L^0,Z),
\end{equation}
where $X_L^0 := W_LX_L+b_L\ones^T$. Again the above is parallelizable across data points.

In the case when the loss function ${\cal L}$ is a squared Frobenius norm, and with a ReLU activation, the update rule~(\ref{eq:Xlast-update-gen}) takes the form
\[
X_{L+1} = \arg \min_{Z \ge 0} \: \|Z-Y\|_F^2 + \lambda_L \|Z-X_L^0\|_F^2,
\]
which can be solved analytically:
\[
X_{L+1}^+ = \max\left(0, \frac{1}{1+\lambda_{L+1}} Y + \frac{\lambda_{L+1}}{1+\lambda_{L+1}}X_L^0 \right).
\]
In the case when the loss function is cross-entropy, and the last layer generates a probability distribution via the probability simplex projection, the above takes the form
\begin{align}\label{eq:cross-entropy-last-layer}
X_{L+1} = \arg \min_{Z} \: &-\Tr Y^T\log Z + \lambda_{L+1} \|Z-X_L^0\|_F^2  \nonumber \\
&Z \ge 0, \;\; Z^T\ones = \ones 
\end{align}
where we use the notation $\log$ in a component-wise fashion. The above can be solved as a set of parallel bisection problems. See Appendix A.

\section{Numerical Experiments} % (fold)
\label{sec:numerical_experiments}
Although lifted models in their own right can be used for supervised learning tasks, their main success so far has been using them to initialize traditional networks. In this section, we examine this and see if the lifted models can generate good initial guesses for standard networks. 
% subsection synthetic_experiments (end)

\subsection{MNIST} % (fold)
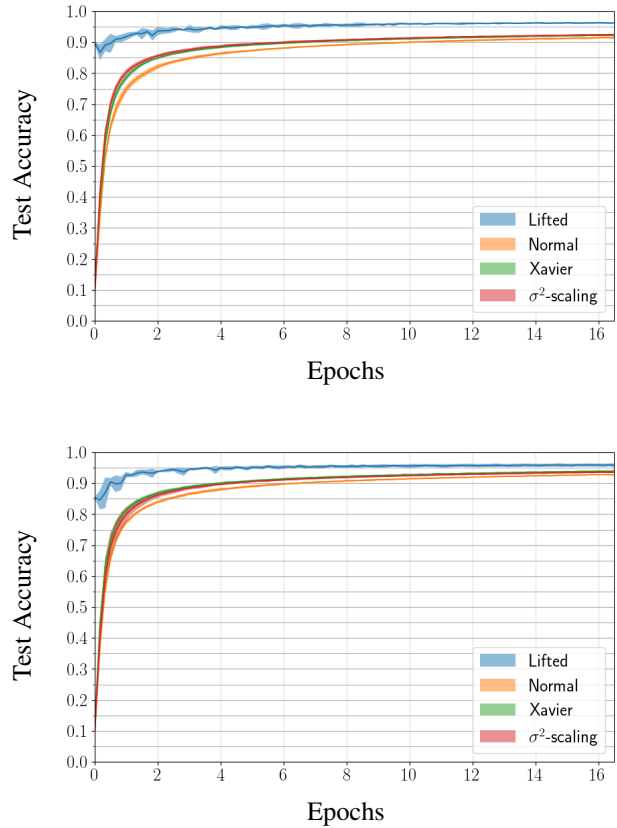
\begin{figure}[h!]
\begin{minipage}{0.5\textwidth}
\begin{tikzpicture}
  \node (img)  {\includegraphics[scale=.35]{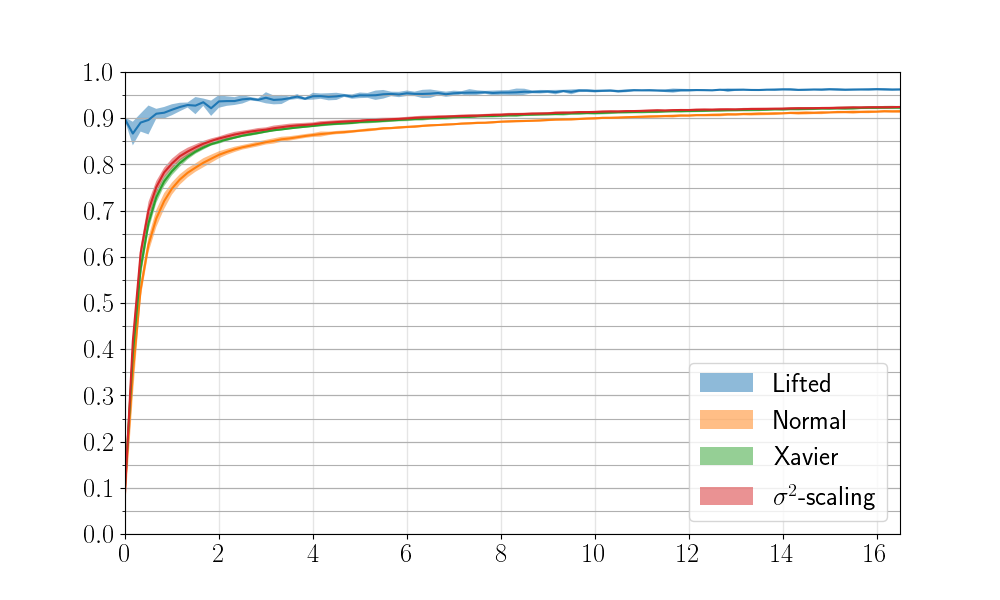}};
  \node[below=of img, node distance=0cm, yshift=1.3cm,font=\color{black}] {Epochs};
  \node[left=of img, node distance=0cm, rotate=90, anchor=center,yshift=-1.3cm,font=\color{black}] {Test Accuracy};
 \end{tikzpicture}
\begin{tikzpicture}
  \node (img)  {\includegraphics[scale=.35]{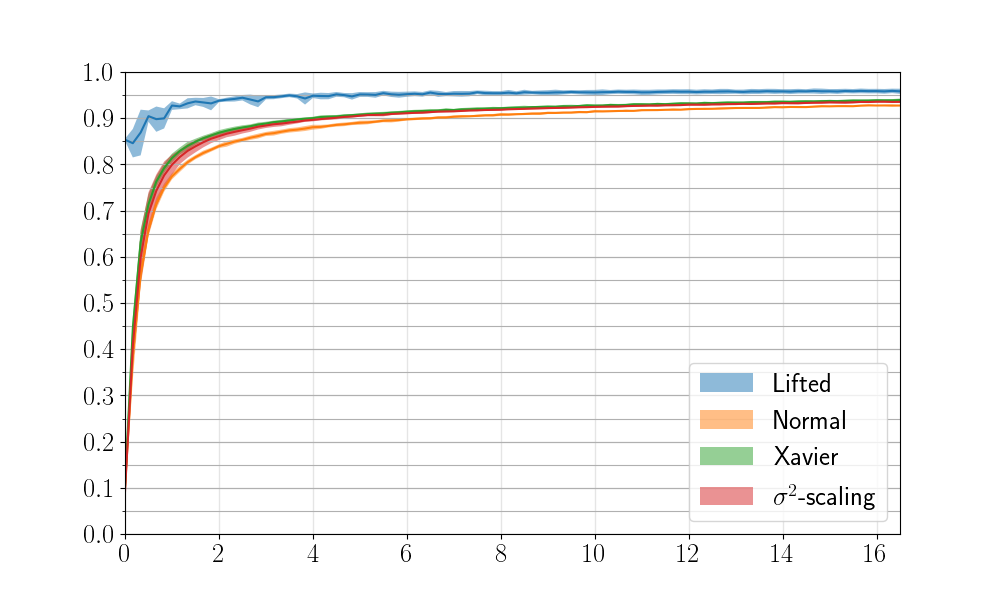}};
  \node[below=of img, node distance=0cm, yshift=1.3cm,font=\color{black}] {Epochs};
  \node[left=of img, node distance=0cm, rotate=90, anchor=center,yshift=-1.3cm,font=\color{black}] {Test Accuracy};
 \end{tikzpicture}
\end{minipage}%
\caption{Plot of test accuracy vs number of training epochs on a held-out validation set during training for two different architectures.  The shaded area on the plots indicated uncertainty to 2 standard deviations across 5 different experiments. The batch size was fixed at $100$ and the learning rate was $\eta = $ \num{1e-5}. \textbf{Top:} One layer neural network with $300$ hidden units and ReLU activation.  \textbf{Bottom:}  Neural network composed of 3 ReLU layers with $500$, $200$, and $100$ hidden units respectively.}
\label{fig:accuracies}
\end{figure}

\label{sub:mnist}
The model described in this paper was compared against a traditional neural network with equivalent architectures on the MNIST dataset \cite{lecun-mnisthandwrittendigit-2010}. For the classification problem, the dataset was split into 60,000 training samples and 10,000 test samples with a softmax cross entropy loss. This is a similar model to the one specified in (\ref{eq:lifted-ridge-lambdas}), with the only difference that the last layer loss is changed from an $\ell_2$ loss to a softmax cross entropy loss as seen in (\ref{eq:example-lifted}). In addition to comparing the models, the weights and biases learned in the augmented neural network were used as initialization parameters for training a standard neural net of the same architecture to compare their performance, both in classification and convergence during training. For all models, ReLU activations were used. $\ell_2$ regularization was used for all layers and the regularization parameters $\rho = 10^{-3}$ were held constant throughout all training procedures. The $\lambda$ parameters for the lifted model were selected using Bayesian Optimization. The lifted model was trained using the block-coordinate descent scheme outlined in Section \ref{sec:block_coordinate_descent_approach}. The standard feedforward networks were trained in Tensorflow using a constant learning rate; reasons for this are highlighted in \citep{DBLP:conf/nips/WilsonRSSR17Marginal}. Table 1 summarizes the accuracy rates for the different architectures for 2 different learning rates. Figure \ref{fig:accuracies} illustrates the test set accuracy versus number of epochs for two different architectures.

\begin{table*}[t]\label{tb:results}
\centering
Learning rate $\eta = $\num{1e-5}
\begin{tabular}{lllllll}
\Xhline{3\arrayrulewidth}
Architecture    & Our Model &NN[Normal] & NN[Xavier] & NN [$\sigma$-scale] & NN [Lifted] \\ \hline
$300$  & $0.898 \pm 0.005$ & $0.915 \pm 0.004$ & $0.9230 \pm 0.005$ & $0.924 \pm 0.003$ & $\mathbf{0.962 \pm 0.003}$ \\
$300-100$  & $0.875 \pm 0.005$ & $0.919 \pm 0.003$ & $0.932 \pm 0.003$ & $0.931 \pm 0.003$ & $\mathbf{0.969 \pm 0.004}$ \\
$500-150$ & $0.865 \pm 0.005$ & $0.927 \pm 0.003$ & $0.936 \pm 0.004$ & $0.935 \pm 0.005$ & $\mathbf{0.970 \pm 0.005}$ \\
$500-200-100$ & $0.853 \pm 0.003$ & $0.927 \pm 0.001$ & $0.939 \pm 0.005$ & $0.935 \pm 0.003$ & $\mathbf{0.958 \pm 0.008}$ \\
$400-200-100-50$ & $0.770 \pm 0.015$ & $0.919 \pm 0.005$ & $\mathbf{0.938 \pm 0.003}$ & $0.936 \pm 0.006$ & $0.919 \pm 0.030$ \\ \hline
\end{tabular}\\
\vspace{3mm}
Learning rate $\eta = $\num{1e-6}
\begin{tabular}{lllllll}
\Xhline{3\arrayrulewidth}
Architecture    & Our Model & NN[Normal] & NN[Xavier] & NN [$\sigma^2$-scale] & NN [Lifted] \\ \hline
$300$   & $0.898 \pm 0.005$ & $0.800 \pm 0.011$ & $0.836 \pm 0.008$ & $0.844 \pm 0.011$ & $\mathbf{0.875 \pm 0.022}$ \\
$300-100$ & $0.875 \pm 0.005$ & $0.792 \pm 0.013$ & $0.838 \pm 0.007$ & $0.842 \pm 0.009$ & $\mathbf{0.899 \pm 0.021}$ \\
$500-150$ & $0.865 \pm 0.005$ & $0.824 \pm 0.007$ & $0.850 \pm 0.004$ & $0.858 \pm 0.002$ & $\mathbf{0.890 \pm 0.018}$ \\
$500-200-100$ & $0.853 \pm 0.003$ & $0.821 \pm 0.017$ & $0.857 \pm 0.007$ & $0.848 \pm 0.011$ & $\mathbf{0.926 \pm 0.053}$ \\
$400-200-100-50$ & $0.770 \pm 0.015$ & $0.751 \pm 0.045$ & $0.838 \pm 0.017$ & $0.815 \pm 0.020$ & $\mathbf{0.959 \pm 0.003}$ \\ \hline
\end{tabular}\\
\caption{Accuracy rate on the test set using different networks with the best result in boldface. The architectures indicate the number of hidden layers and the number of hidden units per layer. NN[$x$] indicates a standard neural network initialized with method $x$: \textit{Normal} for normally distributed intialization of all weight variables with $\mu =  0 $ and $\sigma^2 = 0.1$, \textit{Xavier} for initialization highlighted in \cite{Glorot10understandingthe}, $\sigma^2$-\textit{scale} for variance scaling initialization and \textit{Lifted} for initializing with the weights and biases learned from a lifted NN. All bias variables were initialized to 0.1 except for the Lifted case in which the bias vectors are optimized during pretraining.  The neural networks were trained for 17 epochs using mini-batch gradient descent in Tensorflow \citep{tensorflow2015-whitepaper}.  The lifted model achieves test accuracy as high as 90 \% on MNIST. }
\end{table*}

% \begin{figure}[h!]
% \centering
% \includegraphics[width=0.8\linewidth]{ours_vs_random.png}
% \caption{Both plots show accuracy on a held-out validation set during training for two different architectures. x-axis is the number of ($256$)-batches during training. The left-hand architecture is composed of 4 ReLU layers with $500$, $300$, $150$ and $100$ hidden units respectively. The right-hand architecture is a one layer ReLU network with $1000$ nodes.}
% \label{fig:accuracies}
% \end{figure}

% \begin{figure}[h!]
% \center
% \includegraphics[scale=0.3]{1_300_1e-5_no_titles.png}
% \includegraphics[scale=0.3]{3_500_1e-5_no_titles.png}
% \caption{Plot of test accuracy vs number of training epochs on a held-out validation set during training for two different architectures.  The shaded area on the plots indicated uncertainty to 2 standard deviations across 5 different experiments. The batch size was fixed at $100$. \textbf{Top:} One layer neural network with $300$ hidden units and ReLU activation.  \textbf{Bottom:}  Neural network composed of 3 ReLU layers with $500$, $200$, and $100$ hidden units respectively.}
% \label{fig:accuracies}
% \end{figure}

\begin{remark}
Although our model does not perform as well as the other models on this task, using it as initialization results in increased accuracy for almost all network architectures.
\end{remark}

In particular, in Figure \ref{fig:accuracies} we see that with our initialization, the test accuracy both converges more quickly and to higher values compared with the other initializations: in fact, across all experiments the lifted initlization starts within $90\%$ of its final accuracy. This seems to indicate that the lifted model we train on is a close approximation to a standard feedforward network and our weights learned are already near optimal for these networks. Although after a few passes of the dataset the other models converge, we usually observed a constant gap between the test set accuracy using our initialization versus the others. 
% subsection mnist (end)

\section{Conclusion}
In this work we have proposed a novel model for supervised learning. The key idea behind our method is replacing non-smooth activation functions by smooth penalties in the training problem; we have shown how to do this for general monotonic activation functions. This modifies the multi-layer neural networks optimization problem to a similar problem which we called a lifted neural network. We applied this technique to build a model which we later use as initialization on feedforward neural networks with ReLU activations. Experimental results have shown that the weights of our trained model serve as a good initialization for the parameters of classical neural networks, outperforming neural networks with both random and structured initialization.

%Furthermore, our approach obtains the largest margin of improvement on deeper networks $\textbf{look at this later}$, a very challenging case from an optimization perspective.

\section{Future Work}
Although lifted nets give good results when used as weight initialization for MNIST, they have not extensively been tested on other well known datasets such as CIFAR-10 or other non-image based data sets. The simplest extension of this work will be to apply lifted nets to these different data sets and to different learning tasks such as regression. The lifted framework also easily allows for several extensions and variants that would be very difficult to consider in a standard formulation. This includes handling uncertainty in the data (matrix uncertainty) using principles of robust optimization, optimizing over scale parameters in activation functions, such as the $\alpha$-parameter in leaky-ReLUs, and adding unitary constraints on the $W$ variables. Speedup in a distributed setting is also a point of interest. Additionally, the lifted model can easily be adapted for both convolutional and recurrent neural network architectures.
% \subsection{Software and Data}

% We strongly encourage the publication of software and data with the
% camera-ready version of the paper whenever appropriate. This can be
% done by including a URL in the camera-ready copy. However, do not
% include URLs that reveal your institution or identity in your
% submission for review. Instead, provide an anonymous URL or upload
% the material as ``Supplementary Material'' into the CMT reviewing
% system. Note that reviewers are not required to look a this material
% when writing their review.

% \section*{Acknowledgements}

% \textbf{Do not} include acknowledgements in the initial version of
% the paper submitted for blind review.

% If a paper is accepted, the final camera-ready version can (and
% probably should) include acknowledgements. In this case, please
% place such acknowledgements in an unnumbered section at the
% end of the paper. Typically, this will include thanks to reviewers
% who gave useful comments, to colleagues who contributed to the ideas,
% and to funding agencies and corporate sponsors that provided financial

\bibliography{liftedNNs.bib}
\bibliographystyle{icml2018}

\appendix
\titleformat{\section}{\Large\bfseries}{\thesection}{1em}{}
\gdef\thesection{Appendix \Alph{section}}
\section{Solving for the last layer with cross entropy loss} % (fold)
\label{sec:solving_the_last_layer_with_cross_entropy_loss}
In this section, we consider problem (15), which is of the form
\begin{equation}\label{eq:cross-entropy-last-layer-appendix}
\min_{Z} \: -\Tr Y^T\log Z + \lambda \|Z-X^0\|_F^2 ~:~ Z^T\ones = \ones, \;\; Z \ge 0,
\end{equation}
where we use the notation $\log$ in a component-wise fashion, and $X^0 \in \reals^{p \times m}$ and $Y \in \{0,1\}^{p \times m}$, $Y^T\ones = \ones$, $\lambda>0$ are given. The above can be easily solved by dual matrix bisection. Indeed, the problem can be decomposed across columns of $Y$ (that is, across data points). The problem for a single column has the following form:
\[
p^* := \min_{z} \: -\sum_{i=1}^p y_i \log z_i + \lambda \|z-x^0\|_2^2 ~:~ z \ge 0, \;\; z^T\ones = 1,
\]
where vectors $y \in \{0,1\}^p$, $y^T\ones = 1$ and $x^0 \in \reals^p$ are given. Dualizing the equality constraint,  we obtain a Lagrangian of the form
\[
{\cal L}(z,\nu) = 2\nu + \sum_{i=1}^p \left( z_i^2 - \frac{y_i}{\lambda} \log z_i -2 z_i(\nu + x_i^0) \right) ,
\]
where $\nu$ is a (scalar) dual variable. At the optimum $z^*$, we have 
\[
\forall \: i ~:~ 0 = \frac{1}{2}\frac{\partial {\cal L}(z,\nu)}{\partial z_i}(z^*,\nu) = z_i^* - \frac{y_i}{2\lambda z_i^*} -(\nu + x_i^0),
\]
leading to the unique non-negative solution
\[
z_i^* = \frac{x_i^0+\nu}{2} + \sqrt{\left(\frac{x_i^0+\nu}{2}\right)^2 + \frac{y_i}{2\lambda}}, \;\; i=1,\ldots,p,
\]
where the dual variable $\nu$ is such that $\ones^T z^* = 1$. We can locate such a value $\nu$ by simple bisection. 

The bisection scheme requires initial bounds on $\nu$. For the upper bound, we note that the property $z^* \le \ones$, together with the above optimality condition, implies 
\[
\nu \le  1 - \max_{1 \le i \le p} \: \left(x_i^0 + \frac{y_i}{2\lambda}\right).
\]
For the lower bound, let us first define ${\cal I} := \{ i \::\: y_i \ne 0\}$, $k = |{\cal I}| \le p$. At optimum, we have
\[
\forall \: i \in {\cal I} ~:~ - \log z_i^* \le -\sum_{j \in {\cal I}} y_j \log z_j^* \le p^* \le \theta,
\]
\[
\theta:=-\sum_{i \in {\cal I}} y_i \log z_i^0 + \lambda \|z^0-x^0\|_2^2
\]
where $z^0 \in \reals^p$ is any primal feasible point, for example $z_i^0 = 1/k$ if $i \in {\cal I}$, $0$ otherwise. We obtain 
\[
\forall \: i \in {\cal I}~:~ z_i^* \ge z_{\rm min} :=  e^{-\theta}.
\]
The optimality conditions imply
\[
0 = \frac{1}{2} \sum_{i} \frac{\partial {\cal L}(z^*,\nu)}{\partial z_i} = -\frac{1}{2\lambda} \sum_{i \in {\cal I}} \frac{y_i}{z_i^*} + 1 - \ones^Tx^0 - p\nu 
\]
and therefore:
\[
p\nu =  1 - \ones^Tx^0 -\frac{1}{2\lambda} \sum_{i \in {\cal I}} \frac{y_i}{z_i^*} \ge  1-\ones^Tx^0 -\frac{\ones^Ty}{2\lambda}e^\theta .
\]
To conclude, we have
\[
\frac{1}{p} \left( 1-\ones^Tx^0 -\frac{\ones^Ty}{2\lambda}e^{\theta} \right) =: \underline{\nu} \le \nu \le \overline{\nu} := 1 - \max_{1 \le i \le p} \: \left(x_i^0 + \frac{y_i}{2\lambda}\right).
\]

To solve the original (matrix) problem~(\ref{eq:cross-entropy-last-layer-appendix}), we can process all the columns in parallel (matrix) fashion, updating a \emph{vector} $\nu \in \reals^m$.
% section solving_the_last_layer_with_cross_entropy_loss (end)

\end{document}